\documentclass[conference]{IEEEtran}

\usepackage[utf8]{inputenc}
\usepackage{cite}
\usepackage{amsmath,amssymb,amsfonts}
\usepackage{algorithm}
\usepackage{algorithmic}
\usepackage{graphicx}
\usepackage{textcomp}
\usepackage{xcolor}
\usepackage{booktabs}
\usepackage{multirow}
\usepackage{array}
\usepackage{makecell}
\usepackage{tikz}
\usepackage{url}
\usepackage{balance}
\usepackage[hidelinks]{hyperref}

\hypersetup{
    colorlinks=false,
    pdfborder={0 0 0},
    bookmarksnumbered=true,
    bookmarksopen=true
}
\usetikzlibrary{arrows.meta,positioning,fit,calc,shapes.geometric,shadows.blur}

\definecolor{fameblue}{RGB}{33,92,152}
\definecolor{famegreen}{RGB}{31,135,85}
\definecolor{fameorange}{RGB}{205,112,32}
\definecolor{famepurple}{RGB}{105,78,160}
\definecolor{famered}{RGB}{180,60,55}
\definecolor{famegray}{RGB}{245,247,250}
\definecolor{famedark}{RGB}{35,42,50}

\newcommand{\method}{FAME}
\newcommand{\TopK}{\mathrm{TopK}}
\newcommand{\softmax}{\mathrm{softmax}}
\newcommand{\argminop}{\mathop{\mathrm{arg\,min}}}
\newcommand{\KL}{\mathrm{KL}}

\newcommand{\R}{\mathbb{R}}

\begin{document}

\title{FAME: Forecastability-Aware Mixture of Experts for Heterogeneous Time Series Forecasting}

\author{
\IEEEauthorblockN{Qianyang Li\IEEEauthorrefmark{1}\IEEEauthorrefmark{2}, Xingjun Zhang\IEEEauthorrefmark{1}, Shaoxun Wang\IEEEauthorrefmark{1}, Tao Peng\IEEEauthorrefmark{2} and Jia Wei\IEEEauthorrefmark{3}}

\IEEEauthorblockA{\IEEEauthorrefmark{1}\small\textit{School of Computer Science and Technology, Xi'an Jiaotong University, Xi'an, China}}
\IEEEauthorblockA{\IEEEauthorrefmark{2}\small\textit{Department of Research and Development, Shandong New Beiyang Information Technology Co., Ltd., WeiHai 264200, China}}
\IEEEauthorblockA{\IEEEauthorrefmark{3}\small\textit{Department of Computer Science and Technology, Tsinghua University, Beijing 100084}}
\IEEEauthorblockA{\small Email: \{liqianyang, shaoxunwang\}@stu.xjtu.edu.cn, xjzhang@xjtu.edu.cn}
}

\maketitle

\begin{abstract}
Large-scale retail and industrial forecasting systems contain many heterogeneous time series whose lifecycle, sparsity, volatility, seasonality, spectral patterns, and contextual sensitivity differ substantially. A single forecasting model rarely performs well across all regimes, while dense ensembles increase inference cost and provide limited insight into expert suitability. This paper studies forecastability-aware expert routing: learning how data characteristics determine the suitability of forecasting experts. We propose \method{}, a sparse mixture-of-experts framework that represents each series with a multidimensional forecastability fingerprint, mines expert-suitability targets from validation performance, and trains a cost-aware sparse router to activate a small budgeted set of experts for each series. Using a production-scale vending-machine sales dataset from Shandong New Beiyang (SNBC), where the forecasting component has been integrated into the replenishment-planning pipeline, together with public retail benchmarks, we show that expert suitability varies systematically across data regimes. On the industrial dataset with 5,000+ machines and 60M+ transactions, \method{} Top-2 reduces MSE by 12.4\% over the strongest single expert, LightGBM, while executing 1.92 experts per series on average. The deployed component produces demand forecasts, while inventory-oriented gains are estimated by an offline replay simulator under a fixed replenishment policy rather than by online intervention. The framework turns heterogeneous sales forecasting from heuristic model selection into data mining of forecastability patterns and expert specialization. Code is available at https://github.com/hit636/FAME.
\end{abstract}


\section{Introduction}
Time series forecasting is a core task in data mining and decision support, with applications in retail replenishment, traffic operation, energy management, industrial monitoring, and financial planning~\cite{fildes2022retail}. In many real-world systems, the forecasting target is not a single clean benchmark series but a massive collection of fine-grained sequences. For example, a retail network may require daily forecasts for product--store or product--terminal pairs. Such sequences can be short-lived, intermittent, promotional, stable, seasonal, non-stationary, or strongly affected by context such as location, weather, holiday, and price. This heterogeneity makes universal forecasting difficult.

Existing model families exhibit complementary inductive biases, consistent with the no-free-lunch view~\cite{wolpert1997nfl}. Statistical models, such as exponential smoothing ~\cite{ets}, Prophet ~\cite{prophet}, Croston~\cite{croston}, and TSB~\cite{ets,prophet,croston,tsb}, can be robust for simple seasonal or intermittent patterns. Gradient-boosted decision trees, such as XGBoost~\cite{xgboost}, LightGBM~\cite{lightgbm}, and CatBoost~\cite{catboost}, are strong when lag features, calendar features, product metadata, and contextual covariates are informative. Deep forecasting models, including PatchTST ~\cite{patchtst}, TimeMixer~\cite{timemixer}, TimesNet~\cite{timesnet}, DLinear~\cite{dlinear}, and Mamba-style networks~\cite{smamba,timemachine,mamba4cast}, can model complex temporal structures when enough historical observations are available. However, no family uniformly dominates. A method that is best for sparse cold-start products may be suboptimal for long-history seasonal products; a model that captures abrupt nonlinearity may overfit stable low-volume series.

Most forecasting studies focus on improving the backbone model itself. In large-scale industrial settings, however, the central difficulty is often the heterogeneity of the series rather than the lack of a universally stronger architecture. Stable and seasonal products may be well handled by statistical models, sparse or volatile products often benefit from tree-based methods, and long-history series can be better suited to deep forecasters. This suggests that forecasting performance depends not only on model capacity, but also on the match between data characteristics and model competence. We therefore study forecastability-aware expert routing: learning which forecasting expert is suitable for a series from its observable characteristics.

Industrial sales forecasting provides a natural setting for this problem. Forecastability changes continuously with lifecycle stage, sparsity, intermittency, volatility, trend, seasonality, spectral structure, metadata, and contextual sensitivity. Existing rule-based systems, including our prior USFF framework~\cite{usff}, have shown that assigning different model types to different sales regimes can be useful in production, but they rely on manually defined thresholds and hand-crafted scores. Such rules are difficult to transfer across products, regions, and lifecycle stages, and hard category assignments cannot express that multiple experts may be suitable for the same series. This motivates a learnable, continuous, and cost-aware expert routing framework.

We propose \method{} ({Forecastability-Aware Mixture of Experts}), a sparse expert-routing framework for heterogeneous time series forecasting. \method{} consists of four key components. First, it extracts a forecastability fingerprint from each series, including lifecycle, sparsity, intermittency, volatility, trend, seasonality, spectral, metadata, and context-sensitivity features. Second, it trains a heterogeneous expert pool and evaluates each expert on validation windows to form an expert-loss matrix. Third, it mines hard or soft oracle expert suitability from the loss matrix. Fourth, it trains a sparse router that maps fingerprints to expert probabilities and calls at most $r$ experts under a Top-$r$ budget during inference.

Our contributions are summarized as follows:
\begin{itemize}
    \item We formulate heterogeneous time series forecasting as \emph{forecastability-aware expert routing}, a data mining problem that learns how series characteristics determine expert suitability across regimes, rather than treating model choice as a global benchmark-level decision.
    
    \item We propose \method{}, an oracle-guided sparse mixture-of-experts framework that integrates forecastability fingerprinting, validation-based suitability mining, Top-$r$ budgeted routing, cost-aware regularization, and interpretable expert selection. Unlike dense feature-based averaging, \method{} learns instance-level suitability and executes only active experts, with a lightweight justification of router learning and sparse inference cost.

    \item We conduct production-scale evaluation on the FAME forecasting component deployed in SNBC's replenishment-planning workflow, covering 5,000+ vending machines and 60M+ transactions, together with M5 and Favorita benchmarks~\cite{m5,favorita}. Results show consistent gains over single experts, rule-based routing, stacking, and dense MoE on the industrial dataset, and AutoML-style selection on public benchmarks while using only a small expert subset per series.
\end{itemize}

\section{Related Work}
\subsection{Retail and Industrial Sales Forecasting}
Retail forecasting combines statistical models for seasonal or intermittent series, tree learners for covariate-rich demand, and deep models for long-history nonlinear patterns~\cite{fildes2022retail,m5}. Industrial product--terminal demand is long-tailed: cold-start products, sparse purchases, promotions, weather-sensitive items, and stable staples coexist, motivating adaptive model assignment.

\subsection{Forecastability, FFORMA, and Meta-learning}
Forecastability studies how data properties relate to forecasting difficulty and model performance. FFORMA~\cite{fforma} learns feature-based model-averaging weights, while time-series meta-learning links descriptors to forecasting accuracy~\cite{metalearning,tsfeatures}. \method{} differs from these feature-based meta-learners in three aspects. First, it performs budgeted sparse Top-$r$ activation and executes only the final active experts, whereas FFORMA-style weighting is a dense model-combination scheme. Second, expert suitability is mined from validation-window loss matrices as hard or soft oracle supervision, so the target is series-level expert competence rather than a hand-crafted regime label. Third, routing is driven by explicit forecastability fingerprints, which preserve interpretability and allow expert specialization to be inspected at the series and regime levels.

\subsection{Mixture of Experts, AutoML, and Stacking}
MoE models use gating networks to combine specialized predictors~\cite{moe}. Dense MoE and stacking often execute all experts and offer limited per-series explanation. AutoML systems automate model selection~\cite{autogluon,autogluonts}, but usually optimize global validation performance or return a single selected pipeline. \method{} is different from such global selection: it learns a reusable router from forecastability fingerprints, allows multiple near-optimal experts to share probability mass, and selects at most $r$ active experts under an explicit inference-cost constraint.

\subsection{Foundation Forecasters}
Foundation forecasters such as TimesFM, Chronos, and Moirai~\cite{timesfm,chronos,moirai} and recent sparse time-series MoE models~\cite{timemoe,gatets} provide strong or efficient candidate forecasters. They are complementary to \method{}: a high-cost expert can be added, and the router decides when its cost is justified.

\section{Problem Formulation}
Let $\mathcal{D}=\{(X_i,Y_i,M_i)\}_{i=1}^{N}$ denote a collection of $N$ forecasting instances. $X_i\in\R^{L_i}$ is the historical target sequence, $Y_i\in\R^{H}$ is the future horizon, and $M_i$ contains metadata and contextual variables, such as product category, brand, price, terminal scenario, city, weather, holiday, and promotion indicators. Unlike homogeneous benchmarks, $L_i$ and the statistical patterns of $X_i$ vary substantially.

Let $\mathcal{E}=\{E_1,\ldots,E_M\}$ be a pool of $M$ candidate experts. Each expert outputs
\begin{equation}
    \hat{Y}_{i,m}=E_m(X_i,M_i),\quad m=1,\ldots,M .
\end{equation}
The objective is to learn a router $g_{\phi}$ that maps a fingerprint $z_i$ to an expert suitability vector
\begin{equation}
    p_i=g_{\phi}(z_i),\quad p_i\in[0,1]^M,\quad \sum_{m=1}^{M}p_{i,m}=1 .
\end{equation}
During inference, Top-$r$ is a maximum active-expert budget, not a fixed-cardinality constraint. Let $\mathcal{C}_i^{(r)}=\TopK(p_i,r)$ and $\tilde{p}_{i,m}=p_{i,m}/\sum_{j\in\mathcal{C}_i^{(r)}}p_{i,j}$. The realized active set is
\begin{equation}
\begin{aligned}
\mathcal{K}_i =
\{m\in\mathcal{C}_i^{(r)} \mid
&a_{i,m}=1, \\
&\tilde{p}_{i,m}\ge\delta\},
\quad 1\le |\mathcal{K}_i|\le r\ll M .
\end{aligned}
\end{equation}
Here $a_{i,m}=0$ if an expert lacks required history/covariates or the item is masked by stockout, outage, or unavailability flags, and $1$ otherwise; public benchmarks set it to $1$ for compatible experts. The threshold $\delta$ is tuned on a calibration window from $\{0,0.02,0.05,0.10\}$ and frozen; standard and cost-aware Top-2 use $0.05$ and $0.10$. If pruning empties the set, the highest-probability available candidate is kept.
The sparse forecast is
\begin{equation}
\hat{Y}_i
=
\sum_{m\in\mathcal{K}_i}
\alpha_{i,m}\hat{Y}_{i,m},
\quad
\alpha_{i,m}
=
\frac{p_{i,m}}
{\sum_{j\in\mathcal{K}_i}p_{i,j}}
\end{equation}
The learning problem jointly optimizes accuracy, router fidelity, diversity, and inference cost.

\begin{figure*}[t]
\centering
\includegraphics[width=1.05\textwidth,height=0.31\textheight,keepaspectratio]{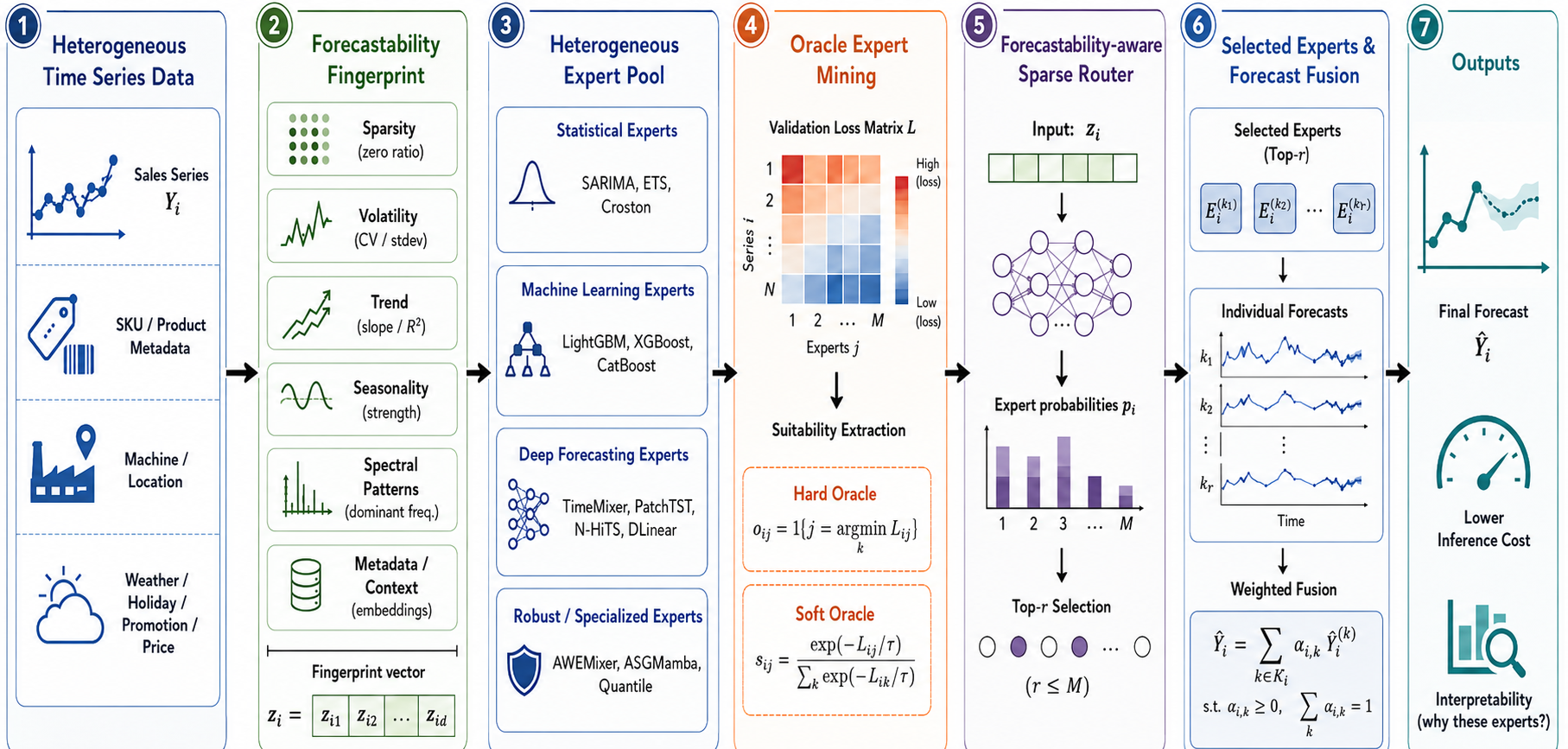}
\caption{Schematic overview of \method{}. Heterogeneous sales series are represented by forecastability fingerprints, expert validation produces suitability signals, and a sparse router selects at most $r$ active experts for forecast fusion. All reported industrial comparisons use the fixed expert pool specified in the baselines and expert-pool subsection.}
\label{fig:fame_arch}
\end{figure*}
\section{The Proposed \method{} Framework}
\subsection{Overview}
\begin{figure*}[t]
\centering
\includegraphics[width=0.98\textwidth,height=0.36\textheight,keepaspectratio]{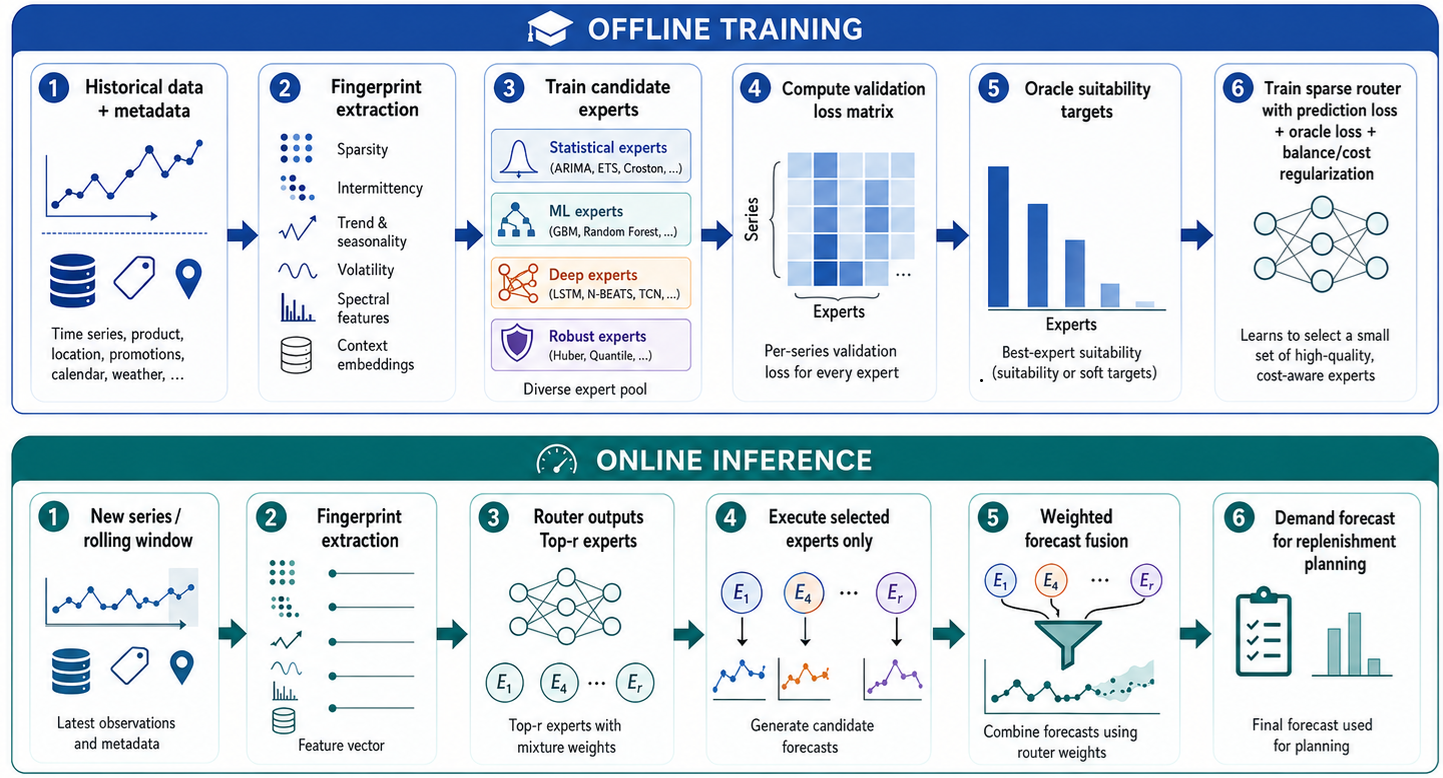}
\caption{Training and inference workflow. Offline training builds the fixed expert pool, validation loss matrix, oracle suitability targets, and sparse router.
Online inference follows Eqs.~(1)--(4): extract the current fingerprint, output expert probabilities, execute only active experts under the Top-$r$ budget, and fuse selected forecasts. Expert names in the schematic are illustrative; reported experiments use the fixed pool in Sec.~V-B.}
\label{fig:fame_detailed}
\end{figure*}

Fig.~\ref{fig:fame_arch} shows the architecture: data alignment, fingerprint extraction, expert validation, oracle mining, and sparse routing. The output is both a forecast and an explanation of expert selection.
Fig.~\ref{fig:fame_detailed} shows the training and inference workflow of \method{}. The complete offline training and online inference procedure is summarized in Algorithm~\ref{alg:fame}, which links the fingerprint extraction, expert validation, oracle suitability mining, router training, and Top-$r$ budgeted sparse forecasting steps.

\subsection{Forecastability Fingerprint}
The fingerprint extractor maps each series to
\begin{equation}
    z_i=[z_i^{\text{life}};z_i^{\text{sparse}};z_i^{\text{vol}};z_i^{\text{trend}};z_i^{\text{season}};z_i^{\text{freq}};z_i^{\text{meta}};z_i^{\text{ctx}}]
\end{equation}

\textbf{Lifecycle features.} We use duration, active days, days since first sale, and days since last sale. These features reflect the available history and product lifecycle.

\textbf{Sparsity and intermittency.} Fine-grained sales series often contain many zero-demand days. We compute
\begin{equation}
    \mathrm{ZeroRatio}_i=\frac{1}{L_i}\sum_{t=1}^{L_i}\mathbb{I}(x_{i,t}=0)
\end{equation}
\begin{equation}
    \mathrm{ADI}_i=\frac{L_i}{\sum_{t=1}^{L_i}\mathbb{I}(x_{i,t}>0)+\epsilon},\quad
    \mathrm{CV}^2_i=\left(\frac{\sigma_i}{\mu_i+\epsilon}\right)^2
\end{equation}
These descriptors distinguish intermittent demand from continuous demand.

\textbf{Volatility and burstiness.} We include coefficient of variation, rolling variance, outlier ratio, and burstiness:
\begin{equation}
    \mathrm{Burstiness}_i=\frac{\sigma_{\Delta x,i}-\mu_{\Delta x,i}}{\sigma_{\Delta x,i}+\mu_{\Delta x,i}+\epsilon}
\end{equation}

\textbf{Trend and seasonality.} Let $X_i=T_i+S_i+R_i$ be an STL-like decomposition~\cite{stl}. Seasonal and trend strengths are
\begin{equation}
    \mathrm{SS}_i=\max\left(0,1-\frac{\mathrm{Var}(R_i)}{\mathrm{Var}(X_i-T_i)+\epsilon}\right)
\end{equation}
\begin{equation}
    \mathrm{TS}_i=\max\left(0,1-\frac{\mathrm{Var}(R_i)}{\mathrm{Var}(X_i-S_i)+\epsilon}\right)
\end{equation}
We also include autocorrelation peaks over business-relevant lags.

\textbf{Spectral features.} Let $F_i(f)=|\mathrm{FFT}(X_i)(f)|^2$ and $\tilde{F}_i(f)=F_i(f)/(\sum_fF_i(f)+\epsilon)$. Spectral entropy and band-energy ratios are
\begin{equation}
    \mathrm{SE}_i=-\sum_f\tilde{F}_i(f)\log(\tilde{F}_i(f)+\epsilon)
\end{equation}
\begin{equation}
    \mathrm{BE}_{i,b}=\sum_{f\in\Omega_b}\tilde{F}_i(f),\quad b\in\{\mathrm{low},\mathrm{mid},\mathrm{high}\}
\end{equation}

\begin{algorithm}[t]
\caption{Training and inference of \method{}}
\label{alg:fame}
\scriptsize
\begin{algorithmic}[1]
\REQUIRE Series collection $\{(X_i,Y_i,M_i)\}_{i=1}^{N}$, expert pool $\mathcal{E}$, routing budget $r$.
\ENSURE Forecasts $\hat{Y}_i$ and routing explanations.
\STATE Preprocess and align target series, metadata, calendar, weather, price, and promotion variables.
\FOR{each series $i$}
    \STATE Extract forecastability fingerprint $z_i$.
\ENDFOR
\FOR{each expert $E_m\in\mathcal{E}$}
    \STATE Train $E_m$ on training windows.
    \STATE Evaluate $E_m$ on oracle-mining windows and store $\ell_{i,m}$.
\ENDFOR
\STATE Mine hard labels $e_i^*$ or soft suitability targets $q_i$ from $\ell_{i,m}$.
\STATE Train router $g_{\phi}$ using $\mathcal{L}_{total}$.
\FOR{each inference series $i$}
    \STATE Compute $p_i=g_{\phi}(z_i)$ and form $\mathcal{C}_i^{(r)}=\TopK(p_i,r)$.
    \STATE Prune unavailable or low-weight candidates to obtain $\mathcal{K}_i$ with $1\le |\mathcal{K}_i|\le r$.
    \STATE Run only active experts $E_m, m\in\mathcal{K}_i$ and fuse them with normalized weights.
\ENDFOR
\RETURN final forecasts and selected-expert explanations.
\end{algorithmic}
\end{algorithm}

\textbf{Metadata and context.} Product category, brand, price band, terminal scenario, city, and other categorical attributes are encoded using embeddings or target statistics. Contextual sensitivity, such as holiday lift and weather sensitivity, is estimated from historical conditional means or correlations.
Table~\ref{tab:fingerprint} summarizes the forecastability fingerprint groups used by \method{} and their representative features.

\begin{table}[t]
\centering
\caption{Forecastability fingerprint groups used by \method{}.}
\label{tab:fingerprint}
\scriptsize
\begin{tabular}{p{1.65cm}p{5.3cm}}
\toprule
Group & Representative features \\
\midrule
Lifecycle & duration, active days, days since first/last sale \\
Sparsity & zero ratio, ADI, nonzero mean, $\mathrm{CV}^2$ \\
Volatility & CV, burstiness, rolling variance, outlier ratio \\
Trend & slope, trend strength, drift, moving-average change \\
Seasonality & seasonal strength, ACF peak, weekly/monthly indicators \\
Spectral & entropy, dominant frequency, low/mid/high band energy \\
Metadata & category, brand, price band, city, terminal scenario \\
Context & weather sensitivity, holiday lift, promotion lift \\
\bottomrule
\end{tabular}
\end{table}

\subsection{Heterogeneous Expert Pool}
The expert pool contains complementary forecasters: statistical experts target short, stable, or intermittent regimes; machine-learning experts exploit contextual covariates; and deep experts capture complex temporal patterns. \method{} is architecture-agnostic, but all industrial comparisons use the same fixed pool so gains come from routing rather than changing candidates. Additional experts such as CatBoost, PatchTST, N-HiTS, Mamba, or foundation forecasters can be added without redesigning the router. Table~\ref{tab:expertpool} summarizes the expert families, representative models, and suitable regimes.

\begin{table}[t]
\centering
\caption{Expert pool used in the reported industrial evaluation. Optional experts can be added by dataset and computation budget.}
\label{tab:expertpool}
\scriptsize
\begin{tabular}{p{1.65cm}p{2.45cm}p{2.85cm}}
\toprule
Family & Examples & Suitable regimes \\
\midrule
Statistical & SARIMA, Holt-Winters/ETS, Prophet, Croston/TSB & short history, stable seasonality, intermittent demand \\
Machine learning & Linear Regression, XGBoost, LightGBM & rich exogenous features, nonlinear covariate effects \\
Deep forecasting & DLinear, TimeMixer, TimesNet & long history, multi-scale patterns, complex dynamics \\
Optional extension & CatBoost, PatchTST, N-HiTS, Mamba/foundation expert & dataset-specific or high-cost regimes \\
\bottomrule
\end{tabular}
\end{table}

\subsection{Oracle Expert Mining}
After training each expert on the training split, we evaluate it on the oracle-mining validation window. Let
\begin{equation}
    \ell_{i,m}=\mathcal{L}(E_m(X_i,M_i),Y_i)
\end{equation}
be the oracle-mining loss of expert $E_m$ on series $i$. The hard oracle is
\begin{equation}
    e_i^*=\argminop_m\ell_{i,m}
\end{equation}
A hard label is simple but ignores cases where multiple experts are similarly good. Therefore, we also define a soft suitability target:
\begin{equation}
    q_{i,m}=\frac{\exp(-\ell_{i,m}/\tau)}{\sum_{j=1}^{M}\exp(-\ell_{i,j}/\tau)}
\end{equation}
where $\tau$ controls softness. To incorporate inference cost, we can use
\begin{equation}
    \tilde{\ell}_{i,m}=\ell_{i,m}+\eta c_m
\end{equation}
where $c_m$ is normalized inference cost. The coefficient $\eta$ constructs cost-aware oracle targets, whereas $\gamma$ in Eq.~(18) regularizes router probabilities during training. We report three operating points: \method{}-Acc prioritizes accuracy, \method{} Top-2 is the calibrated balanced setting, and \method{}-CostAware further reduces batch cost.

\subsection{Forecastability-Aware Sparse Router}
The router predicts expert probabilities from fingerprints:
\begin{equation}
    p_i=g_{\phi}(z_i)=\softmax(\mathrm{MLP}_{\phi}(z_i))
\end{equation}
Top-$r$ routing forms $\mathcal{C}_i^{(r)}=\TopK(p_i,r)$ and prunes unavailable or low-weight candidates to obtain $\mathcal{K}_i$ with $1\le |\mathcal{K}_i|\le r$. Sparse routing mitigates the computational bottleneck, as evaluating the full dense ensemble incurs prohibitive latency. During router training, experts are pretrained and frozen. We compute $\mathcal{L}_{pred}$ using the differentiable soft mixture $\tilde{Y}_i=\sum_{m=1}^{M} p_{i,m}\hat{Y}_{i,m}$; inference applies the Top-$r$ budget and renormalizes active weights. Numerical features use robust train-split standardization, categorical metadata use embeddings or smoothed target statistics from training windows only, and the router is a two-layer MLP with ReLU, dropout, and temperature-calibrated softmax.

\textbf{Train--inference consistency.} The router is supervised by the full soft suitability distribution because oracle-mining losses provide graded near-optimal expert signals. Budgeted Top-$r$ projection and pruning introduce a truncation gap, controlled when most oracle mass lies in the active set. We therefore monitor Top-$r$ oracle recall and compare sparse-aware variants in Sec.~\ref{sec:ablation}. Straight-through Top-$r$, Gumbel-Top-$r$, and SparseMAX did not outperform the calibrated soft-KL router because they produced less stable expert usage under noisy validation losses.

\subsection{Training Objective}
The overall objective combines forecasting, router supervision, load balancing, and cost regularization:
\begin{equation}
    \mathcal{L}_{total}=\mathcal{L}_{pred}+\lambda\mathcal{L}_{router}+\beta\mathcal{L}_{balance}+\gamma\mathcal{L}_{cost}.
\end{equation}
The prediction loss during router training uses the differentiable soft mixture $\tilde{Y}_i$:
\begin{equation}
    \mathcal{L}_{pred}=\frac{1}{N}\sum_{i=1}^{N}\mathcal{L}(\tilde{Y}_i,Y_i).
\end{equation}
The router supervision loss is either cross-entropy with hard labels or KL divergence with soft oracle targets:
\begin{equation}
    \mathcal{L}_{router}=\frac{1}{N}\sum_{i=1}^{N}\KL(q_i\Vert p_i).
\end{equation}
Load balancing prevents expert collapse:
\begin{equation}
    \bar{p}_m=\frac{1}{N}\sum_i p_{i,m},\quad
    \mathcal{L}_{balance}=M\sum_{m=1}^{M}\bar{p}_m^2
\end{equation}
The cost term is
\begin{equation}
    \mathcal{L}_{cost}=\frac{1}{N}\sum_i\sum_{m=1}^{M} p_{i,m}c_m
\end{equation}

\subsection{Design Justification}
The following analysis provides a theoretical intuition for the router objective and sparse execution design. It explains why oracle imitation, KL supervision, and cost-aware sparse routing are reasonable design choices.
\subsubsection{Routing Regret}
Let $\ell_{i,m}$ be bounded in $[0,B]$. Let the oracle distribution be $q_i$ and the learned router distribution be $p_i$. Define the expected routing loss under distribution $p_i$ as $R_i(p_i)=\sum_{m=1}^{M}p_{i,m}\ell_{i,m}$. Then
\begin{equation}
    |R_i(p_i)-R_i(q_i)|\le B\|p_i-q_i\|_1
\end{equation}
This follows directly from Holder's inequality. Thus, if the router approximates the oracle suitability distribution, its expected loss approaches the oracle-guided routing loss. By Pinsker's inequality,
\begin{equation}
    \|p_i-q_i\|_1\le \sqrt{2\KL(q_i\Vert p_i)}
\end{equation}
which justifies the KL supervision loss. Minimizing $\mathcal{L}_{router}$ reduces a bound on routing regret.

\subsubsection{Cost Reduction by Sparse Routing}
Let $c_m$ be the inference cost of expert $E_m$. A full ensemble costs $C_{full}=\sum_{m=1}^{M}c_m$, while budgeted sparse routing costs
\begin{equation}
    C_i=\sum_{m\in\mathcal{K}_i}c_m, \quad |\mathcal{K}_i|\le r .
\end{equation}
For comparable experts, $C_i/C_{full}\le r/M$ and is smaller when pruning activates fewer than $r$ experts; for costly deep experts, the cost term further discourages unnecessary calls.

\subsubsection{Fingerprint-Based Routing Rationale}
We route experts with explicit forecastability fingerprints instead of relying only on latent sequence embeddings. In industrial sales data, many series are short, sparse, or intermittent, making purely learned routing features unstable. Descriptors such as zero ratio, ADI, seasonal strength, and spectral entropy provide robust signals about expert suitability and make the routing process inspectable. Learned representations can be concatenated when enough history is available, but the fingerprint remains the primary interface for both routing and explanation.

\section{Experimental Evaluation}

\subsection{Industrial Production Dataset}

The primary dataset is an anonymized production-scale vending-machine sales log from Shandong New Beiyang (SNBC), whose forecasting service is integrated into the replenishment-planning pipeline. It records daily demand from 5,000+ vending machines, about 5,000 products, and over 60 million transactions across multiple replenishment cycles. The forecasting unit is a product--terminal--day series with historical demand, terminal scenario, city, product ID, category, brand, price, holidays, weather, and availability flags. Zero-demand days are retained when the product is active and available.

The logs follow chronological 70\%/10\%/20\% training, validation, and test splits. The 10\% validation window is further split into equal oracle-mining and router-calibration parts: the former builds the expert-loss matrix and labels, while the latter tunes $\delta$, $\tau$, $\gamma$, early stopping, and model selection. The final 20\% window is held out. Series with fewer than 15 active days are removed; missing dates are filled with zero sales and active-status flags; discontinued products are masked after their last sale; and fingerprints use only the look-back segment. Days with outage, replenishment suspension, or stockout/unavailable flags are excluded from expert-loss construction and replay. Table~\ref{tab:deployment_profile} summarizes the production deployment profile of the industrial dataset and forecasting system.

\begin{table}[tb]
\centering
\caption{Production forecasting deployment and offline replay protocol.}
\label{tab:deployment_profile}
\scriptsize
\setlength{\tabcolsep}{3pt}
\begin{tabular}{p{2.1cm}p{5.2cm}}
\toprule
Item & Description \\
\midrule
Company & Shandong New Beiyang (SNBC) vending-machine operation \\
Scale & 5,000+ vending machines, about 5,000 products, 60M+ transactions \\
Forecast unit & product--terminal--day demand series \\
Forecast horizon & 14 days for replenishment planning \\
Context & city, terminal scenario, product attributes, price, weather, holidays, availability flags \\
Deployment mode & scheduled batch forecasting with periodic expert and router refresh \\
Business usage & deployed demand forecasts; inventory effects estimated by fixed-policy offline replay, not online A/B testing  \\
\bottomrule
\end{tabular}
\end{table}

The 14-day horizon matches the replenishment cycle. Each expert uses the same history and supported covariates. Statistical experts are trained per series or compatible group; tree models are global learners with lag, rolling, calendar, price, metadata, and context features; deep models use fixed look-back windows. In production, the daily job runs after transaction aggregation and before planning; forecasts feed downstream services, while inventory results are reported from offline replay. Router features are historical only to prevent leakage.

\subsection{Baselines and Expert Pool}

We compare single experts, dense ensembles, and routing baselines. The industrial expert pool is fixed to the same $M=10$ candidates used by all routing and ensemble methods: SARIMA, Holt-Winters/ETS, Prophet~\cite{prophet}, Croston/TSB, Linear Regression, XGBoost~\cite{xgboost}, LightGBM~\cite{lightgbm}, DLinear~\cite{dlinear}, TimeMixer~\cite{timemixer}, and TimesNet~\cite{timesnet}. Routing baselines include USFF~\cite{usff}, cluster-then-forecast, stacking, AutoML-style selection, FFORMA-style dense weighting, dense soft MoE, FAME variants, and a validation-selected oracle reference. Stacking uses ridge regression over validation forecasts; AutoML-style selection trains a LightGBM classifier on fingerprints.

Table~\ref{tab:main_result} reports representative single experts for readability; all ensemble and routing methods use the fixed pool above, so improvements are attributed to routing rather than changing candidates.

\subsection{Implementation Setup}

Hyperparameters are selected on the router-calibration split, with final settings in Table~\ref{tab:hyper}. Neural experts and the router are trained with three seeds; we report mean test performance. Deterministic experts use fixed chronological splits and seeds. Industrial comparisons use paired Wilcoxon signed-rank tests over product--terminal test windows, with grouped bootstrap checks used as auxiliary evidence. Runtime is normalized wall-clock batch inference time against LightGBM.

\begin{table}[b]
\centering
\caption{Representative hyperparameters used in production evaluation.}
\label{tab:hyper}
\scriptsize
\setlength{\tabcolsep}{2pt}
\begin{tabular}{p{1.4cm}p{5.7cm}}
\toprule
Model & Main settings \\
\midrule
SARIMA & seasonal period 7; order selected on train/validation grid \\
Prophet & weekly seasonality; holiday regressors; changepoint prior tuned \\
Holt-Winters & additive trend; multiplicative or additive seasonality selected by validation \\
LightGBM & Tweedie objective; learning rate 0.015; num leaves 256; subsample 0.65 \\
XGBoost & squared-error regression; max depth 10; learning rate 0.015; subsample 0.65 \\
TimeMixer & look-back 49; horizon 14; batch size 128; early stopping on validation \\
DLinear & look-back 49; horizon 14; batch size 128; RevIN-style normalization \\
TimesNet & $d_{model}=256$; 4 layers; 8 heads; look-back 49; horizon 14 \\
Router & two-layer MLP; hidden size 128; dropout 0.1; Top-$r$ budgeted selection \\
\bottomrule
\end{tabular}
\end{table}
Experiments run on dual Intel(R) Xeon(R) Gold 5218 CPUs, 256 GB RAM, and two NVIDIA RTX 4090 GPUs, using Python 3.10, PyTorch 2.1, LightGBM 4.1, XGBoost 2.0, and CUDA 12.1. Public benchmark code covers preprocessing, rolling splits, fingerprint extraction, expert training, validation-loss mining, and router training. Industrial data cannot be released for confidentiality; we provide the schema, configurations, and protocol, with fixed splits before model selection.

\subsection{Metrics}
\begin{table*}[t]
\centering
\caption{Industrial production results. Single-expert rows are representative; routing/ensemble rows use the fixed ten-expert pool. ``Exec.'' is realized active experts after pruning. Cost is normalized by LightGBM; the oracle reference is diagnostic. Negative reductions indicate worse MSE than LightGBM.}
\label{tab:main_result}
\scriptsize
\setlength{\tabcolsep}{4pt}
\begin{tabular}{lcccccc}
\toprule
Method & MSE & MAE & WAPE & Exec. & Norm. Cost & MSE Reduction vs. LightGBM \\
\midrule
SARIMA & 4.798 & 4.044 & 0.418 & 1.0 & 0.8 & -203.8\% \\
Holt-Winters & 4.307 & 3.624 & 0.384 & 1.0 & 0.8 & -172.7\% \\
Linear Regression & 1.798 & 1.486 & 0.173 & 1.0 & 0.6 & -13.9\% \\
XGBoost & 1.714 & 1.438 & 0.168 & 1.0 & 1.1 & -8.5\% \\
LightGBM & 1.579 & 1.348 & 0.157 & 1.0 & 1.0 & 0.0\% \\
TimeMixer & 1.845 & 1.635 & 0.181 & 1.0 & 3.6 & -16.8\% \\
TimesNet & 1.833 & 1.610 & 0.179 & 1.0 & 4.1 & -16.1\% \\
Uniform Ensemble & 1.566 & 1.302 & 0.151 & 10.0 & 16.4 & 0.8\% \\
FFORMA-style Weighting & 1.514 & 1.255 & 0.146 & 10.0 & 16.4 & 4.1\% \\
Stacking Ensemble & 1.472 & 1.218 & 0.142 & 10.0 & 17.0 & 6.8\% \\
Rule-based USFF & 1.489 & 1.239 & 0.144 & 1.0 & 1.2 & 5.7\% \\
Cluster-then-Forecast & 1.523 & 1.274 & 0.148 & 1.0 & 1.2 & 3.6\% \\
Dense Soft MoE & 1.438 & 1.191 & 0.139 & 10.0 & 16.5 & 8.9\% \\
\textbf{\method{} Top-1} & \textbf{1.421} & \textbf{1.174} & \textbf{0.137} & 1.0 & 1.3 & \textbf{10.0\%} \\
\textbf{\method{} Top-2} & \textbf{1.384} & \textbf{1.143} & \textbf{0.133} & 1.92 & 2.4 & \textbf{12.4\%} \\
\method{}-CostAware & 1.397 & 1.155 & 0.135 & 1.54 & 1.8 & 11.5\% \\
Validation-selected Oracle Reference & 1.326 & 1.097 & 0.128 & 1.0 & N/A & 16.0\% \\
\bottomrule
\end{tabular}
\end{table*}
We report MSE, MAE, WAPE, stock-aware cost, normalized inference cost, and realized active experts. Runtime is measured end-to-end after common aggregation: shared features are counted once, while expert-specific preprocessing, inference, and fusion are counted only for active experts. For forecast $\hat{y}$ and target $y$, the stock-aware cost is
\begin{equation}
\mathcal{L}_{\mathrm{stock}}
=
\omega_u \max(y-\hat{y},0)
+
\omega_o \max(\hat{y}-y,0)
\end{equation}
where $\omega_u>\omega_o$ gives higher penalty to stockout risk than to redundant inventory or expiration loss. Router quality is evaluated by Top-1 accuracy, Top-2/Top-3 oracle recall, usage entropy, and oracle gap.

\subsection{Feature Thresholds and Expert Hyperparameters}
Although \method{} uses continuous fingerprints, Table~\ref{tab:thresholds} reports coarse diagnostic regimes for interpretation and USFF-style comparison; they are not routing rules.

\begin{table}[t]
\centering
\caption{Coarse regime thresholds used for diagnostic reporting.}
\label{tab:thresholds}
\scriptsize
\setlength{\tabcolsep}{3pt}
\begin{tabular}{lll}
\toprule
Dimension & Regime & Definition \\
\midrule
Duration & short / medium / long & $[15,60)$ / $[60,270)$ / $\ge270$ active days \\
Volatility & low / high & CV $<0.5$ / CV $\ge0.5$ \\
Seasonality & weak / strong & seasonal strength $<0.5$ / $\ge0.5$ \\
Sparsity & low / high & zero ratio $<0.4$ / $\ge0.4$ \\
Intermittency & low / high & ADI $<1.32$ / $\ge1.32$ \\
Spectral & concentrated / diffuse & spectral entropy below / above median \\
\bottomrule
\end{tabular}
\end{table}

\section{Main Results}
\subsection{Overall Industrial Results}
Table~\ref{tab:main_result} reports the main results on the production-scale industrial dataset. LightGBM is the strongest single expert among the individual models. This confirms the strength of global tree models for retail demand because they exploit lag, rolling, calendar, product, terminal, and context features. However, rule-based USFF already improves over the best single expert by assigning different regimes to different experts. The learned \method{} router further improves accuracy while preserving sparse inference. Dense soft MoE is accurate but expensive because all experts are executed. The default \method{} Top-2 gives the best balanced accuracy--cost point, while \method{}-CostAware trades a small amount of accuracy for fewer active experts and lower batch cost.

\subsection{Router Quality and Expert Specialization}
Table~\ref{tab:regime_detail} summarizes representative regime-level results. Statistical experts are strong for short stable sequences. LightGBM frequently achieves the best performance in many medium-history nonlinear regimes. XGBoost is competitive when volatility is high but seasonality is weak. TimeMixer is preferred for long-history sequences with strong periodicity. These patterns justify routing: model suitability is associated with forecastability fingerprints.

Table~\ref{tab:router_quality} directly evaluates whether the router learns expert suitability. Top-$s$ recall measures whether the validation-selected expert appears in the first $s$ router-ranked experts; oracle gap is the relative MSE gap to the validation-selected oracle reference. \method{} improves oracle recall over AutoML-style selection, explaining why Top-2 budget routing gives a better accuracy--cost trade-off than hard Top-1 routing. \method{} Top-1 and Top-2 share the same learned router ranking; they differ only in the maximum routing budget and the resulting active set after pruning, which explains the same ranking metrics but different oracle gaps.

\begin{table*}[t]
\centering
\caption{Regime-level specialization on the industrial test set. Counts are anonymized and rounded for confidentiality.}
\label{tab:regime_detail}
\footnotesize
\setlength{\tabcolsep}{4pt}
\renewcommand{\arraystretch}{1.08}
\begin{tabular}{ccccrrlcc}
\toprule
Cat. & Duration & Vol. & Seas. & \#Series & Txn. (M) & Best Expert & MSE & MAE \\
\midrule
1  & Short  & Low  & Weak   & 18.2K & 2.1 & SARIMA              & 1.134 & 0.714 \\
2  & Short  & Low  & Strong & 13.6K & 1.6 & Holt-Winters/SARIMA & 1.254 & 0.929 \\
3  & Short  & High & Weak   & 24.2K & 4.2 & TimesNet            & 1.781 & 1.623 \\
4  & Short  & High & Strong & 21.1K & 3.8 & LightGBM            & 1.553 & 1.220 \\
5  & Medium & Low  & Weak   & 28.7K & 5.0 & LightGBM            & 1.203 & 1.092 \\
6  & Medium & Low  & Strong & 25.0K & 4.7 & XGBoost             & 1.816 & 1.534 \\
7  & Medium & High & Weak   & 31.4K & 6.3 & LightGBM            & 1.535 & 1.377 \\
8  & Medium & High & Strong & 29.6K & 5.9 & LightGBM            & 1.570 & 1.440 \\
9  & Long   & Low  & Weak   & 35.9K & 8.7 & LightGBM            & 1.445 & 1.126 \\
10 & Long   & Low  & Strong & 30.9K & 7.5 & TimeMixer           & 1.490 & 1.242 \\
11 & Long   & High & Weak   & 41.1K & 5.4 & LightGBM            & 1.235 & 1.063 \\
12 & Long   & High & Strong & 38.9K & 5.0 & TimeMixer           & 1.502 & 1.131 \\
\bottomrule
\end{tabular}
\end{table*}

\begin{table}[t]
\centering
\caption{Router quality on industrial test windows. Ent. denotes expert-usage entropy.}
\label{tab:router_quality}
\scriptsize
\setlength{\tabcolsep}{3pt}
\begin{tabular}{lccccc}
\toprule
Router & Top-1 Acc. & Top-2 Rec. & Top-3 Rec. & Gap & Ent. \\
\midrule
AutoML selector & 0.42 & 0.57 & 0.70 & 0.147 & 1.21 \\
\method{} Top-1 & 0.51 & 0.68 & 0.81 & 0.079 & 1.86 \\
\method{} Top-2 & 0.51 & 0.68 & 0.81 & 0.062 & 1.86 \\
\method{}-CostAware & 0.48 & 0.63 & 0.77 & 0.071 & 1.61 \\
\bottomrule
\end{tabular}
\end{table}

\subsection{Public Retail Benchmarks}
We also evaluate on M5 and Favorita. M5 uses item--store sales with calendar events, SNAP, prices, and hierarchy~\cite{m5}. Favorita uses active store--item sales with holidays, oil price, and metadata~\cite{favorita}; histories shorter than 60 days are filtered. Both use chronological 28-day validation and test horizons. Fingerprints use only look-back windows. The public pool matches the industrial pool when compatible; metadata-specific inputs are replaced by dataset-compatible variants, $a_{i,m}=1$ for compatible experts, and the same $\delta$ grid is used. Table~\ref{tab:public} shows that \method{} outperforms routing and ensemble baselines under sparse inference.

\begin{table}[t]
\centering
\caption{Public retail benchmark results. Lower WAPE and sMAPE are better.}
\label{tab:public}
\scriptsize
\setlength{\tabcolsep}{2pt}
\begin{tabular}{lcccc}
\toprule
Method & \multicolumn{2}{c}{M5} & \multicolumn{2}{c}{Favorita} \\
 & WAPE & sMAPE & WAPE & sMAPE \\
\midrule
Best single expert & 0.259 & 0.184 & 0.187 & 0.135 \\
Rule-based USFF & 0.251 & 0.178 & 0.181 & 0.131 \\
AutoML selector & 0.245 & 0.174 & 0.176 & 0.127 \\
Stacking ensemble & 0.242 & 0.172 & 0.175 & 0.126 \\
Dense soft MoE & 0.237 & 0.168 & 0.170 & 0.123 \\
\method{} Top-1 & 0.239 & 0.170 & 0.171 & 0.124 \\
\textbf{\method{} Top-2} & \textbf{0.234} & \textbf{0.166} & \textbf{0.168} & \textbf{0.121} \\
Validation-selected reference & 0.226 & 0.160 & 0.162 & 0.117 \\
\bottomrule
\end{tabular}
\end{table}

\textbf{Foundation expert check.} Adding Chronos as an optional expensive expert improves Top-2 WAPE from 0.234 to 0.230 on M5 and from 0.168 to 0.165 on Favorita, while invoking it for only 8.6\% and 10.9\% of series. This supports using foundation models as specialized high-cost experts rather than always-on replacements.

\section{Ablation and Cost Analysis}
\label{sec:ablation}
\subsection{Ablation Study}
Table~\ref{tab:ablation} evaluates the contribution of major components. Removing sparsity and intermittency features hurts performance on low-volume and intermittent products. Removing metadata/context features leads to the most pronounced performance drop because product category, price band, terminal scenario, weather, and holiday sensitivity are highly informative for vending-machine demand. Removing oracle supervision reduces the router to a weak black-box classifier and increases the oracle gap. The \method{}-Acc row removes the router cost regularizer: it slightly improves error but increases cost because the router selects expensive experts more often and prunes fewer candidates. The active count is at most two because Top-2 is a maximum budget, not a fixed-cardinality constraint.

\begin{table}[t]
\centering
\caption{Ablation on the industrial dataset. Exec. is the realized average number of active experts under the Top-2 budget; cost is normalized by LightGBM. The default row is the calibration-selected operating point, while \method{}-Acc prioritizes error at higher cost.}
\label{tab:ablation}
\scriptsize
\setlength{\tabcolsep}{3pt}
\begin{tabular}{lcccc}
\toprule
Variant & MSE & MAE & Exec. & Cost \\
\midrule
Default \method{} Top-2 & 1.384 & 1.143 & 1.92 & 2.4 \\
w/o sparsity features & 1.427 & 1.180 & 1.93 & 2.4 \\
w/o seasonality features & 1.419 & 1.171 & 1.92 & 2.4 \\
w/o spectral features & 1.410 & 1.164 & 1.91 & 2.4 \\
w/o metadata/context & 1.446 & 1.199 & 1.94 & 2.4 \\
w/o oracle loss & 1.468 & 1.219 & 1.98 & 2.5 \\
w/o balance loss & 1.406 & 1.162 & 1.66 & 2.1 \\
\method{}-Acc Top-2 & 1.372 & 1.135 & 2.00 & 4.3 \\
\bottomrule
\end{tabular}
\end{table}

\textbf{Router and hyperparameter sensitivity.} Straight-through Top-$r$, Gumbel-Top-$r$, and SparseMAX obtain MSE values of 1.392, 1.397, and 1.401, compared with 1.384 for the calibrated soft-KL router. Sweeping $\tau\in\{0.1,0.3,1.0\}$ gives MSE $\{1.397,1.384,1.398\}$. Sweeping the cost coefficient shows the expected accuracy--cost trade-off: removing cost regularization gives MSE/cost $(1.372,4.3)$, the default calibrated setting gives $(1.384,2.4)$, and stronger cost regularization gives $(1.397,1.8)$ and $(1.424,1.4)$.

\subsection{Cost--Accuracy Trade-off}
Figure-style trade-off is summarized in Table~\ref{tab:cost}. Full ensembles slightly improve over single experts but require executing all experts. The default \method{} Top-2 captures most of the oracle margin with fewer than two active experts on average because low-weight or unavailable candidates can be skipped. \method{}-CostAware further lowers cost while retaining most accuracy gains.

\begin{table}[t]
\centering
\caption{Cost--accuracy comparison. Exec. is realized average active experts. Batch time is normalized by LightGBM on the same server.}
\label{tab:cost}
\scriptsize
\setlength{\tabcolsep}{3pt}
\begin{tabular}{lcccc}
\toprule
Method & MSE & MAE & Exec. & Batch Cost \\
\midrule
LightGBM & 1.579 & 1.348 & 1.0 & 1.0 \\
Rule-USFF & 1.489 & 1.239 & 1.0 & 1.2 \\
Uniform ensemble & 1.566 & 1.302 & 10.0 & 16.4 \\
Dense soft MoE & 1.438 & 1.191 & 10.0 & 16.5 \\
\method{} Top-1 & 1.421 & 1.174 & 1.0 & 1.3 \\
\method{} Top-2 & 1.384 & 1.143 & 1.92 & 2.4 \\
\method{}-CostAware & 1.397 & 1.155 & 1.54 & 1.8 \\
\bottomrule
\end{tabular}
\end{table}

\subsection{Business-Oriented Offline Simulation}
The deployed SNBC pipeline uses \method{} forecasts as inputs to replenishment planning, but the inventory outcomes reported here are not online A/B results. They are obtained from an offline replay in which only the demand forecast is replaced, while the order-up-to policy, availability masks, capacity constraints, and replenishment logic are kept identical across methods. For product--terminal $i$ at day $t$, the target level is
\begin{align}
S_{i,t} &= \widehat{D}_{i,t:t+H} + z_{0.95}\widehat{\sigma}_{i,t}\sqrt{L+H}, \\
Q_{i,t} &= \max(0, S_{i,t} - I^{\mathrm{pos}}_{i,t}),
\end{align}
where $L=1$ is the lead time, $H=14$ is the planning horizon, $I^{\mathrm{pos}}_{i,t}$ is inventory position, and $z_{0.95}$ sets the 95\% service target. The replay advances inventory chronologically using the same availability masks, capacity constraints, and replenishment policy for all forecasting methods. A stockout event is counted when post-demand inventory is zero while realized demand is positive, and overstock exposure is the remaining inventory above the target level. The asymmetric loss uses $\omega_u=3\omega_o$. Compared with the best single expert, \method{} Top-2 reduces stock-aware loss by 13.8\%. Compared with the production rule-based USFF selector, it reduces simulated stockout events by 6.4 \% and simulated overstock exposure by 3.7 \%. \method{}-CostAware retains most of this benefit while reducing daily batch cost by 25.0\% relative to the default Top-2 setting.

\section{Interpretability and Deployment}

\subsection{Feature Importance and Expert Usage}
We use permutation importance and binned expert usage to inspect the trained router. Table~\ref{tab:importance} shows that sparsity, seasonality, volatility, duration, and context features dominate routing. Intermittent series favor statistical experts; seasonal or long-history series favor SARIMA, ETS, or TimeMixer; context-sensitive products favor tree models.

\begin{table}[!b]
\centering
\caption{Router feature importance on the production dataset.}
\label{tab:importance}
\scriptsize
\setlength{\tabcolsep}{3pt}
\begin{tabular}{lcc}
\toprule
Feature group & Importance & Typical routing effect \\
\midrule
Zero ratio / ADI & 0.21 & intermittent/statistical experts \\
Seasonal strength / ACF peak & 0.18 & SARIMA, ETS, TimeMixer \\
CV / burstiness & 0.16 & LightGBM, XGBoost, robust experts \\
Duration / active days & 0.15 & deep experts for long histories \\
Holiday sensitivity & 0.11 & tree models with context \\
Spectral entropy & 0.10 & robust or tree experts \\
Price band / category & 0.05 & metadata-aware tree models \\
City / terminal scenario & 0.04 & context-aware experts \\
\bottomrule
\end{tabular}
\end{table}

\subsection{Representative Routing Cases}
Table~\ref{tab:case} gives representative production cases. The selected experts match interpretable forecastability patterns, including intermittent cold-start demand, reliable seasonality, context-driven nonlinear effects, and noisy local fluctuations.

\begin{table}[t]
\centering
\caption{Representative routing cases from production logs.}
\label{tab:case}
\scriptsize
\setlength{\tabcolsep}{2pt}
\begin{tabular}{p{1.2cm}p{2.2cm}p{2.5cm}p{1.4cm}}
\toprule
Case & Fingerprint pattern & Top experts & Explanation \\
\midrule
A & high zero ratio, short duration & Croston/LightGBM & intermittent cold-start demand \\
B & long history, strong weekly seasonality & TimeMixer/SARIMA & reliable periodic structure \\
C & high CV, high holiday lift & LightGBM/XGBoost & context-driven nonlinear effects \\
D & high spectral entropy, weak seasonality & XGBoost/robust expert & noisy local fluctuations \\
\bottomrule
\end{tabular}
\end{table}

\subsection{Leakage Control and Reproducibility}
The system separates expert training, oracle mining, router calibration, and test evaluation. At each forecast origin, fingerprints use only look-back observations and covariates. Categorical statistics are fitted on training data and frozen for validation/test; holiday, weather, and promotion sensitivity use historical deviations only. Suitability labels come from the oracle-mining subwindow, while $\delta$, $\tau$, $\gamma$, early stopping, and model selection use the later calibration subwindow. No test oracle is used. Logged availability, outage, suspension, and stockout masks are used only when known before or at replay time. Censored days are excluded from expert-loss construction and demand-error metrics, while simulated stockouts in replay are counted.

Table~\ref{tab:repro} summarizes the reproducibility and leakage-control protocol. Public scripts reproduce the M5 and Favorita experiments; the confidential industrial data are replaced by schema, configurations, and routing protocol. Potential failure cases, such as unseen regimes, noisy oracle labels, or over-penalized costly experts, are monitored through oracle error, routing stability, loss matrices, usage entropy, and the Top-$r$/cost-coefficient trade-off.

\begin{table}[!t]
\centering
\caption{Reproducibility and leakage-control settings.}
\label{tab:repro}
\scriptsize
\setlength{\tabcolsep}{3pt}
\begin{tabular}{p{2.1cm}p{5.3cm}}
\toprule
Item & Setting \\
\midrule
Prediction unit & product--terminal--day series \\
Split & chronological 70\%/10\%/20\% by calendar time \\
Look-back & rolling windows ending before forecast origin \\
Horizon & 14 days for industrial replenishment \\
Fingerprint & computed only from look-back history \\
Oracle labels & mined from first validation subwindow only \\
Router calibration & $\delta$, $\tau$, $\gamma$ tuned on later validation subwindow \\
Expert cost & measured end-to-end wall-clock time normalized by LightGBM \\
Router model & two-layer MLP over standardized fingerprint features \\
Feature scaling & robust z-score or quantile scaling by train split \\
Censored days & historical stockout/outage/unavailable days excluded from loss; simulated replay stockouts counted \\
Public code & public benchmark scripts and industrial protocol/schema released \\
\bottomrule
\end{tabular}
\end{table}

\begin{table}[t]
\centering
\caption{Paired comparison on industrial test windows.}
\label{tab:sig}
\scriptsize
\setlength{\tabcolsep}{2pt}
\begin{tabular}{lccccc}
\toprule
Comparison & MSE $\Delta$ & MAE $\Delta$ & Improved & Test & $p$-value \\
\midrule
\method{} Top-2 vs. LightGBM & 12.4\% & 15.2\% & 64.8\% & Wilcoxon & $<10^{-4}$ \\
\method{} Top-2 vs. Rule-USFF & 7.1\% & 7.7\% & 59.3\% & Wilcoxon & $<10^{-4}$ \\
\method{} Top-2 vs. Stacking & 6.0\% & 6.2\% & 57.1\% & Wilcoxon & $<10^{-3}$ \\
\method{} Top-2 vs. Dense MoE & 3.8\% & 4.0\% & 54.6\% & Wilcoxon & $0.006$ \\
Validation reference vs. \method{} Top-2 & 4.2\% & 4.0\% & -- & diagnostic & -- \\
\bottomrule
\end{tabular}
\end{table}

\subsection{Statistical Significance and Oracle Gap}
We compare \method{} Top-2 with the strongest non-oracle baselines using paired Wilcoxon signed-rank tests over product--terminal windows, supported by grouped bootstrap checks. Table~\ref{tab:sig} reports relative gains and improved-window fractions. The validation reference is diagnostic and not an inference-time baseline.

\subsection{Production Forecasting Deployment and Monitoring}
\method{} has been deployed at SNBC as the demand-forecasting component of the replenishment-planning workflow. Daily transaction logs are aggregated into product--terminal demand series and aligned with price, calendar, weather, and location context. During scheduled batch inference, the router extracts the current fingerprint, selects a Top-$r$ candidate set, prunes unavailable or low-weight experts, renormalizes active weights, and writes daily demand forecasts to downstream planning services.

The deployment claim refers to this operational forecasting component. Reported stockout and overstock results are not online A/B outcomes; they are fixed-policy offline replay estimates in which only the demand forecast is replaced. Monitoring tracks fingerprint drift, expert-usage entropy, oracle gap, routing stability, and expert-pool coverage for router refresh and emerging-regime detection.

\section{Conclusion}
This paper presented \method{}, a forecastability-aware sparse mixture-of-experts framework for heterogeneous time series forecasting. Instead of designing another universal forecasting backbone, \method{} mines the relationship between data characteristics and expert suitability. It extracts interpretable forecastability fingerprints, derives oracle suitability from validation behavior, and trains a sparse cost-aware router for Top-$r$ budgeted expert selection. Experiments on production-scale vending-machine data with more than 5,000 machines and over 60 million transactions show that expert suitability varies systematically across regimes, that learned sparse routing outperforms single experts, rule-based USFF, stacking, and dense ensembles under a favorable cost--accuracy trade-off, and that routing decisions provide actionable forecastability insights for replenishment planning; the inventory gains should be interpreted as replay-based estimates under a fixed policy. The proposed framework turns heterogeneous sales forecasting from a heuristic model-selection problem into a data mining problem of discovering forecastability patterns and expert specialization.

\end{document}